\pdfoutput=1
\documentclass[sigconf, nonacm]{acmart}

\usepackage{graphicx} 
\usepackage{subfigure}
\usepackage{balance}
\AtBeginDocument{%
  \providecommand\BibTeX{{%
    \normalfont B\kern-0.5em{\scshape i\kern-0.25em b}\kern-0.8em\TeX}}}

\copyrightyear{2020}
\acmYear{2020}
\setcopyright{acmcopyright}\acmConference[MM '20]{Proceedings of the 28th ACM International Conference on Multimedia}{October 12--16, 2020}{Seattle, WA, USA}
\acmBooktitle{Proceedings of the 28th ACM International Conference on Multimedia (MM '20), October 12--16, 2020, Seattle, WA, USA}
\acmPrice{15.00}
\acmDOI{10.1145/3394171.3413772}
\acmISBN{978-1-4503-7988-5/20/10}
\begin{document}
\fancyhead{}
\title{Towards Accuracy-Fairness Paradox: Adversarial Example-based Data Augmentation for Visual Debiasing}




\author{Yi Zhang$^{1,2}$, Jitao Sang$^{1,2}$}
\affiliation{%
  \institution{$^{1}$School of Computer and Information Technology \& Beijing Key Lab of Traffic Data Analysis and Mining, Beijing Jiaotong University, Beijing, China}
}
\affiliation{%
  \institution{$^{2}$Peng Cheng Laboratory, ShenZhen, China}
}

\email{yicheung@bjtu.edu.cn, jtsang@bjtu.edu.cn}


\begin{abstract}
Machine learning fairness concerns about the biases towards certain protected or sensitive group of people when addressing the target tasks. This paper studies the debiasing problem in the context of image classification tasks. Our data analysis on facial attribute recognition demonstrates (1) the attribution of model bias from imbalanced training data distribution and (2) the potential of adversarial examples in balancing data distribution. We are thus motivated to employ adversarial example to augment the training data for visual debiasing. Specifically, to ensure the adversarial generalization as well as cross-task transferability, we propose to couple the operations of target task classifier training, bias task classifier training, and adversarial example generation. The generated adversarial examples supplement the target task training dataset via balancing the distribution over bias variables in an online fashion. Results on simulated and real-world debiasing experiments demonstrate the effectiveness of the proposed solution in simultaneously improving model accuracy and fairness. Preliminary experiment on few-shot learning further shows the potential of adversarial attack-based pseudo sample generation as alternative solution to make up for the training data lackage.
\end{abstract}



\maketitle

\section{Introduction}
\begin{figure}[ht]
  \centering
  \includegraphics[width=0.98\linewidth]{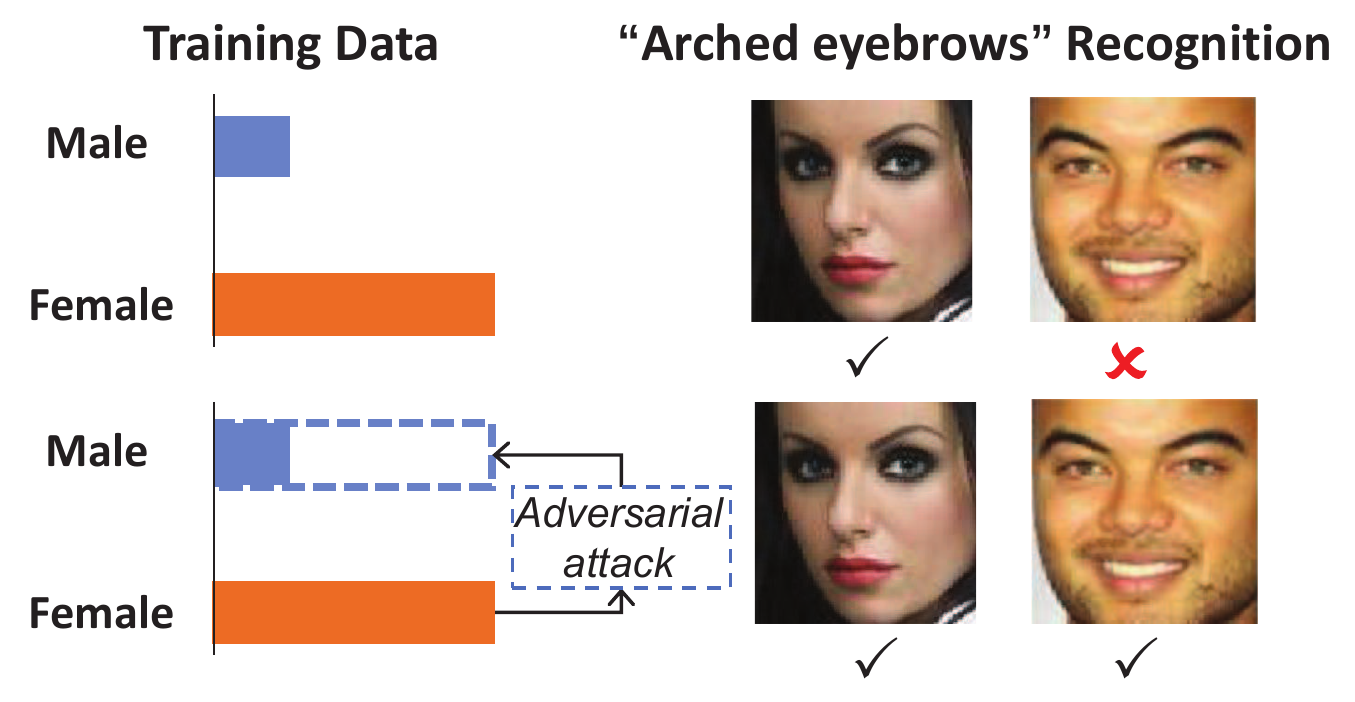}
  \setlength{\abovecaptionskip}{3pt}
  \caption{We propose debiasing method that uses adversarial attacks to balance the data distribution.}
  \label{fig1}
\end{figure}

The last few years have witnessed human-level AI in view of accuracy for tasks like image recognition, speech processing and reading comprehension~\cite{he2016deep,rajpurkar2016squad}. 
The increasing deployment of AI systems drives researchers to pay attention to other criteria beyond accuracy, such as security~\cite{zhang2018emotion}, privacy~\cite{xu2018trust,cciftcci2017reliable} and fairness, 
which are critical for large-scale robust real-world applications~\cite{escalante2018explainable}. Among these criteria, fairness concerns about the potential discrimination of model towards protected or sensitive groups (e.g., gender, skin color). 
A biased model can lead to unintended social consequences such as in online advertisement, banking and criminal justice~\cite{sweeney2013discrimination,hardt2016equality,angwin2016machine}, e.g., the popular COMPAS algorithm for recidivism prediction was found biased against 
black inmates and prone to make unfair sentencing decisions~\cite{angwin2016machine}. 

Machine learning fairness study involves with two types of variables: target and biased variables. In the COMPAS example, target variable is the recidivism label and biased variable is the skin color. The existing model debiasing attempts can be roughly categorized according to the intervention period to make the prediction of target variables independent of the bias variables: 
(1) Pre-processing debiasing modifies the training dataset before learning the model. Typical pre-processing solutions resort to sampling or reweighing the training samples~\cite{drummond2003c4,zhou2005training,kamiran2012data} to make target task model not learning the correlation between target and bias variables. (2) In-processing debiasing modifies the standard model learning during the training process~\cite{d2017conscientious,bellamy2018ai}.  Adv debiasing is the most popular in-processing method, which adversarially optimizes between the target and bias tasks with goal to extract fair representation that only contributes to the target task~\cite{wadsworth2018achieving,beutel2017data,ganin2014unsupervised}. (3) Post-processing debiasing changes output labels to force fairness after model inference, which enhances outcomes to unprivileged groups and weakens outcomes to privileged groups. Post-processing solution is usually employed only when it is difficult to modify training data or training process ~\cite{bellamy2018ai, berk2017convex, mehrabi2019survey}.

Among the three model debiasing categories, in-processing and post-processing both compromise between fairness and accuracy by either imposing additional constraint or explicitly alternating model outputs. Pre-processing enjoys advantage to address the intrinsic attribution of model bias in imbalanced training data distribution, which is widely recognized and further validated in our data analysis. However, conventional pre-processing debiasing solutions employing sampling and reweighing strategies either fails to make full use of the training data or only guarantees superficial data balance. This leads to the reported decreased accuracy in previous debiasing studies. 

The problem thus transfers to how to supplement and balance the data distribution without damaging the original target task learning.  Adversarial example recently draws massive attention regarding model robustness and AI system security~\cite{szegedy2013intriguing}. Other than the endless game between adversarial attack and defense, an interesting line of related studies observed that model trained on adversarial examples discovered useful features unrecognizable by human~\cite{ilyas2019adversarial}. This inspires us to employ adversarial examples to supplement and balance the data distribution for model debiasing. Specifically, given target task (e.g., predicting the facial attribute of ``arched eyebrow'' from image) with imbalanced training set over bias variables (e.g., much more female than male samples with ``arched eyebrow'' annotation), adversarial attack is conducted to alter the bias variable and construct a balanced training set (illustrated in Figure~\ref{fig1}). To guarantee the adversarial generalization and cross-task transferability, we propose an online coupled adversarial attack mechanism to iteratively optimize between target task training, bias task training and adversarial example generation. This can be actually regarded as an debiasing attempt between pre-processing and in-processing, which favorably combines their both advantages. 
We summarize our main contributions as follows:
\begin{itemize}
\item We propose to employ adversarial examples to balance training data distribution in the way of data augmentation. Simultaneously improved accuracy and fairness are validated from simulated and real-world debiasing evaluation.
\item We provide an online coupled adversarial example generation mechanism, which ensures both the adversarial generalization and cross-task transferability.
\item We explore the potential of adversarial examples as supplementary samples, which provides alternative perspective of employing adversarial attack and opens up possibility to addressing data lackage issue from new ways. 
\end{itemize}

\section{Data Analysis and Motivation Justification}
In this section, we first examine the attribution of model bias in image classification tasks, and then analyze the potential and challenges of using adversarial examples to address the model bias problem.

\subsection{Bias Attribution Analysis}\label{sec2.1}
\noindent\textbf{Classification bias definition.}\hspace{2mm} 
People can be divided into different groups based on their social attributes such as \emph{gender, skin color}.  Model bias refers to the unfair and unequal treatment of individuals with certain social attribute, e.g., correlating \emph{arched eyebrow} more with \emph{female} than with \emph{male} in task of facial attribute prediction. 
We use \emph{equality of opportunity} in this work to evaluate model bias, where ideal fairness outcome requires that each group of people have an equal 
opportunity of correct classification. In terms of classification tasks, model bias is formally defined as follows: 

\textsc{Definition 1} (\textsc{Classification Bias}). 
\emph{We derive the bias of target task class $t \in \mathbb{T}$ in a classifier in terms of group difference as: } 
\begin{equation} \label {eqn1}
\begin{aligned}
&bias(\theta,t) \\=&{|P(\hat{t} = t|b = 0,t^* = t) - P(\hat{t} = t|b = 1,t^* = t)|}
 \end{aligned}
\end{equation}
where $\theta$ indicates the parameter of the examined classifier, $\hat{t} \in \mathbb{T}$ denotes the predicted target variable of classifier, $t^* \in \mathbb{T}$ represents the ground-truth target variable like \emph{arched eyebrow}, and $b$ represents the bias variables such as \emph{gender, skin color}.  
Smaller $bias(\theta,t)$ means that the classifier tends not to be affected by the bias variables when making predictions. The sum of the bias of all target variables to measure the overall bias for model $bias(\theta) = \sum_t bias(\theta,t)$.

\noindent\textbf{Attribution in imbalanced data distribution.}\hspace{2mm}  With the definition of model bias, we then use the CelebA dataset~\cite{liu2018large} to examine its attribution in data distribution. Specifically, using the 34 facial attributes~\footnote{~\small{Within the 40 annotated atttributes of CelebA, we waived the ones not related to face (e.g., wearing hat) or essentially belonging to certain gender (e.g., mustache). This leaves 18 attributes, among which the gender attribute is selected as the bias variable. Regarding each of the remaining 17 facial attributes, the bias data distribution is very different for sample sets w/ and w/o this attribute. To facilitate data analysis and later experimental evaluation, we consider each facial attribute in two target tasks (e.g., attribute of arched eyebrow involve with two target variables of arched eyebrow and non-arched eyebrow), which leads to finally $17*2=34$ target tasks.}} 
as target variables to predict and the gender as bias variable, we trained facial attributes classifier and calculated their corresponding gender bias according to Eqn.~\eqref{eqn1}. 

\begin{figure}[t]
  \centering
  \includegraphics[width=0.95\linewidth]{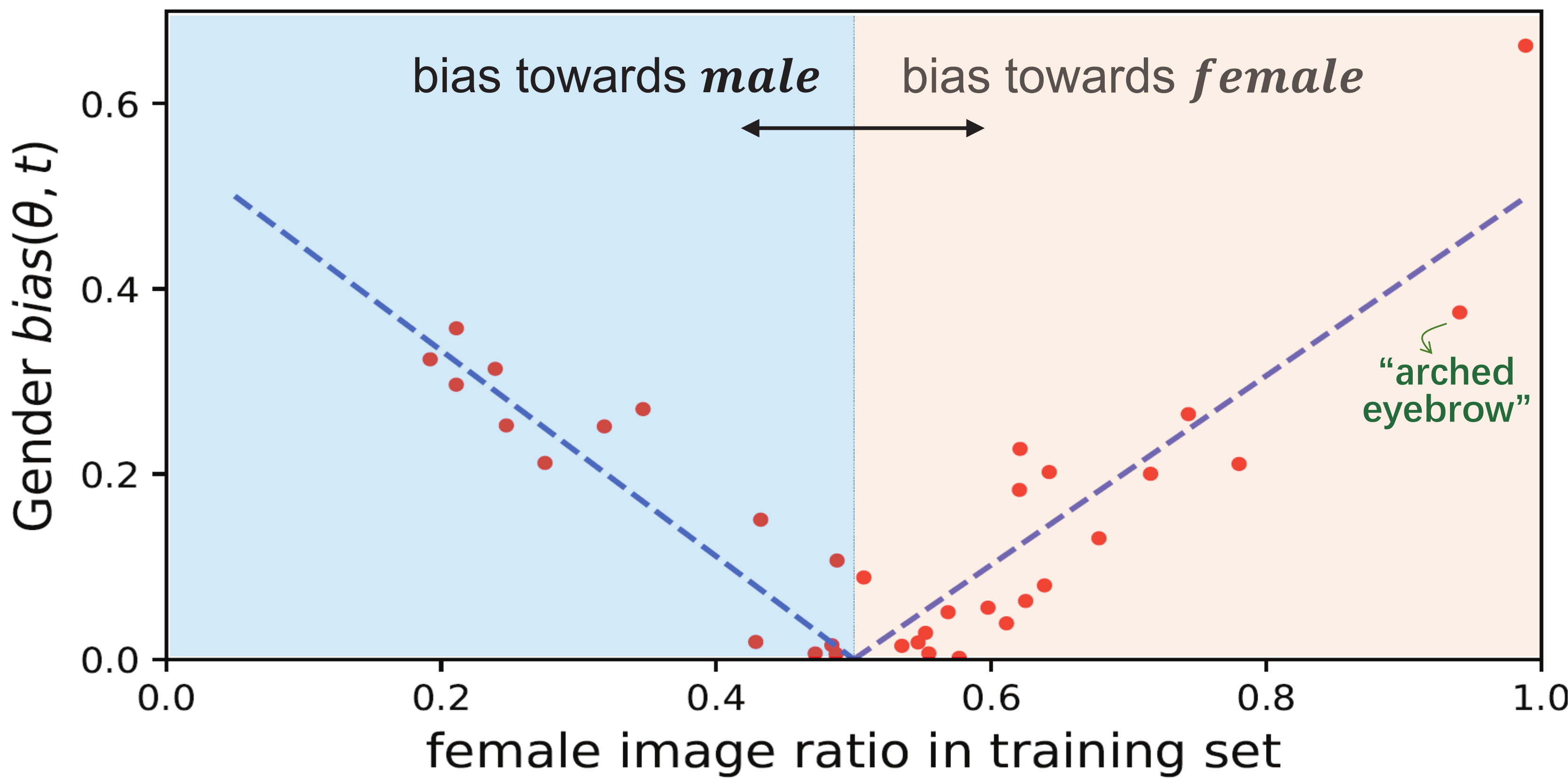}
  \setlength{\belowcaptionskip}{-2mm}
  \caption{Model bias v.s. imbalanced data distribution. $x$-axis denotes the female ratio of total people with certain facial attribute in the training set, and $y$-axis denotes the model bias over gender in predictions testing set.}
  \label{fig2}
\end{figure}

Figure~\ref{fig2} shows the calculated model bias ($y$-axis) for different facial attributes and their corresponding female training image ratio ($x$-axis). It is easy to find the strong correlation between model bias and imbalanced data distribution: for facial attributes with a larger ratio of \emph{female} in training set ($>0.5$ in the $x$-axis), \emph{female} images are more easily correctly classified than \emph{male} images, and vice versa for \emph{male} ($<0.5$ in the $x$-axis). For example, there are more female training images for facial attribute of ``arched eyebrows'', and the corresponding classifier is observed to derive more correct prediction for female images, while male images with \emph{arched eyebrows} are likely to be incorrectly predicted. The observation suggests that the classifier learns the correlation between facial attribute and gender from the imbalanced data, and thus utilizes the gender bias variable for target variable prediction. It well validates the motivation of the previous debiasing attempts via pre-processing to balance training data distribution, so that the learned model will not utilize the bias variables for target task prediction. 

\subsection{The Potential of Adversarial Example in Balancing Data Distribution}\label{sec2.2}
\noindent\textbf{Feasibility of adversarial example for attack class training.}\hspace{2mm}  The above analysis attributes model bias to the imbalanced data distribution over bias variables. However, with the training data reflecting the real-world distribution, it is difficult to explicitly collect more samples with minority-bias variables~\footnote{~\small{We use "minority-bias variable" to denote the value of bias variable with less samples, e.g., male in gender bias.}} (e.g., it is indeed rare to see many males with ``arched eyebrows''). Conventional pre-processing debiasing solutions resort to down-sampling and up-sampling~\cite{drummond2003c4,zhou2005training,kamiran2012data}, which either fails to make full use of the training data or only guarantees superficial data balance.

It is observed from recent studies that adversarial examples contain generalized features of the attack class~\cite{ilyas2019adversarial}, i.e., model trained solely on adversarial examples with attack labels performs well on unmodified real testing data. Inspired by this, we are interested in employing adversarial attack to generate pseudo samples for the minority-bias variables, e.g., with the target task of predicting ``arched eyebrows'', adversarial perturbation is added to attack \emph{female} into \emph{male} images. In this way, the generated pseudo samples can be seen as augmented data for data distribution balancing and thus contribute to model debiasing.

We conducted preliminary experiment to justify the feasibility of adversarial examples in switching gender labels and generalizing to original real samples. Specifically, we first trained binary gender classifier $g_{ori}$ with original face images from the CelebA dataset, and then employed I-FGSM~\cite{kurakin2016adversarial} to attack each original image to its adversarial image with opposite gender label. Denoting the original image set as $\mathcal{X}_{ori}$ and the attacked adversarial image set as $\mathcal{X}_{adv}$, we constructed the following two training datasets: (1) \emph{Hard switch}: original image set $\mathcal{X}_{ori}$ with manually switched gender labels~\footnote{~\small{E.g., training label is set as \emph{male} if the ground-truth label is \emph{female}.}}; (2) \emph{ADV switch}: adversarial image set $\mathcal{X}_{adv}$ with attacked gender labels~\footnote{~\small{E.g., if a \emph{male} image is attacked as \emph{female}, its training label is set as \emph{female}.}}.  

\begin{figure}[t]
  \centering
  \includegraphics[width=\linewidth]{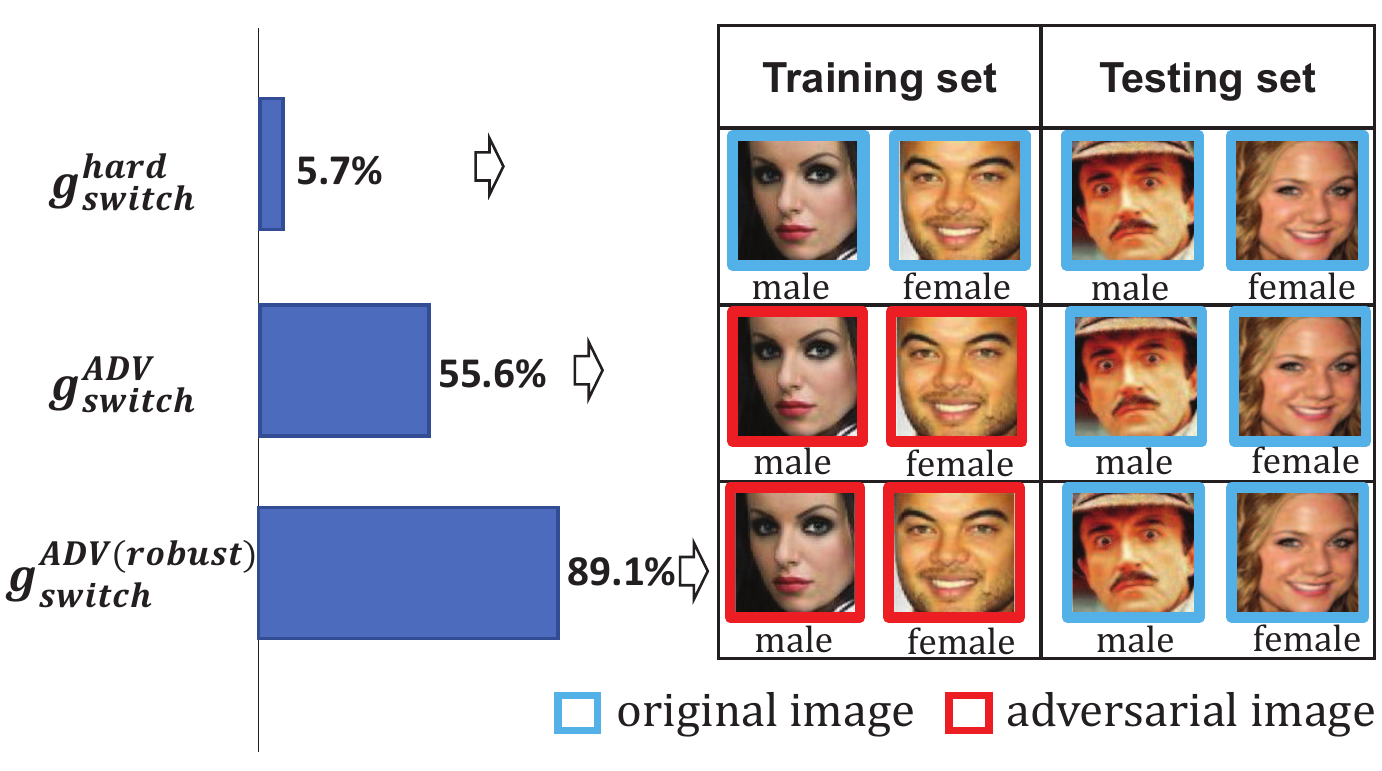}
    \setlength{\abovecaptionskip}{3pt}
  \setlength{\belowcaptionskip}{-1mm}
  \caption{Gender classification accuracy for different training settings with switched labels.}
  \label{fig3}
\end{figure}

We utilized the above two datasets to train gender classifiers $g_{switch}^{hard}$ and $g_{switch}^{ADV}$ respectively. Figure~\ref{fig3} (top 2 rows) shows their classification accuracy on the original image testing set.  It is easy to understand the extremely poor performance of $g_{switch}^{hard}$ as the manually switched labels make the image-label correlation exactly the opposite between the training and testing sets. While by replacing the original images with adversarial attacked images, gender classification accuracy increases from $5.7\%$ to $55.6\%$, verifying that adversarial examples contain useful information about the attack class and have potential to generalize to original real data.

\noindent\textbf{Stronger adversarial examples contributing to improved attack class training.}\hspace{2mm}  However, the accuracy of $55.6\%$ indicates that the adversarial examples are still far from adequate to directly replace real attack-class samples.  Actually, studies have found that the generalization of the adversarial example to attack class largely depends on its attack strength~\cite{nakkiran2019discussion}. That is, adversarial examples fooling more classifiers likely to contain more generalized features to attack class. 

Following this spirit, we expect that a more robust bias classifier can generate stronger adversarial examples generalizing well to attack class. Therefore, we first conducted adversarial training on $g_{ori}$ to improve its robustness and acquired the robust classifier $g_{robust}$, and then employed I-FGSM to attack this robust classifier $g_{robust}$ to derive new training set \emph{ADV switch (robust)} with attacked gender label. The learned gender classifier from this new training set is denoted as $g_{switch}^{ADV (robust)}$, whose classification accuracy is shown in the bottom of Figure~\ref{fig3}. The significant increase from $55.6\%$ to $89.1\%$ demonstrates the superior generalization potential of adversarial examples from robust models, which motivates us to design more robust bias classifiers in generating adversarial examples for data augmentation.

\subsection{Cross-task Transferability}\label{sec2.3}
The above analysis verifies that adversarial examples hold some generalization to attack class in the task of bias classification. 
For model debiasing, two tasks are involved, i.e., the bias task like gender classification and the target task like ``arched eyebrows'' prediction. 
Therefore, in addition to the adversarial generalization within bias task, it is desirable the adversarial example maintains its generalization ability to original real data during training the target task. 
We refer the adversarial examples' capability in maintaining attack class information from bias tasks to target tasks as \emph{cross-task transferability}. 

This subsection examines the cross-task transferability by utilizing adversarial examples for data augmentation-based visual debiasing. We first introduced one straightforward way: the derived adversarial examples $\mathcal{X}_{adv}$ from the previous subsection 
is added into the original training image set to train the target facial attribute classifiers. Specifically, the resultant training set $\mathcal{X}_{augment}$ consists of $\mathcal{X}_{adv}$ and the original images, 
and the target classifier is realized with VGG-16 containing feature extractor $f(\cdot)$ and classification module $h_t(\cdot)$. To examine cross-task transferability, 
using target classifier's feature extractor $f(\cdot)$, we trained additional bias classification module $h_b(\cdot)_{generalize}$ based on $\mathcal{X}_{adv}$ with attacked gender labels, 
and calculated the gender classification accuracy of $\mathcal{X}_{org}$ on bias classifier $\{f(\cdot);h_b(\cdot)_{generalize}\}$. 
Since $\{f(\cdot);h_b(\cdot)_{generalize}\}$ shares the feature extractor $f(\cdot)$ for target task prediction, if original image can be correctly classified by this bias classifier, 
we consider $\mathcal{X}_{adv}$ contains information about the attack class and possesses cross-task transferability to some extent. 
\begin{figure}[t]
  \centering
  \includegraphics[width=0.93\linewidth]{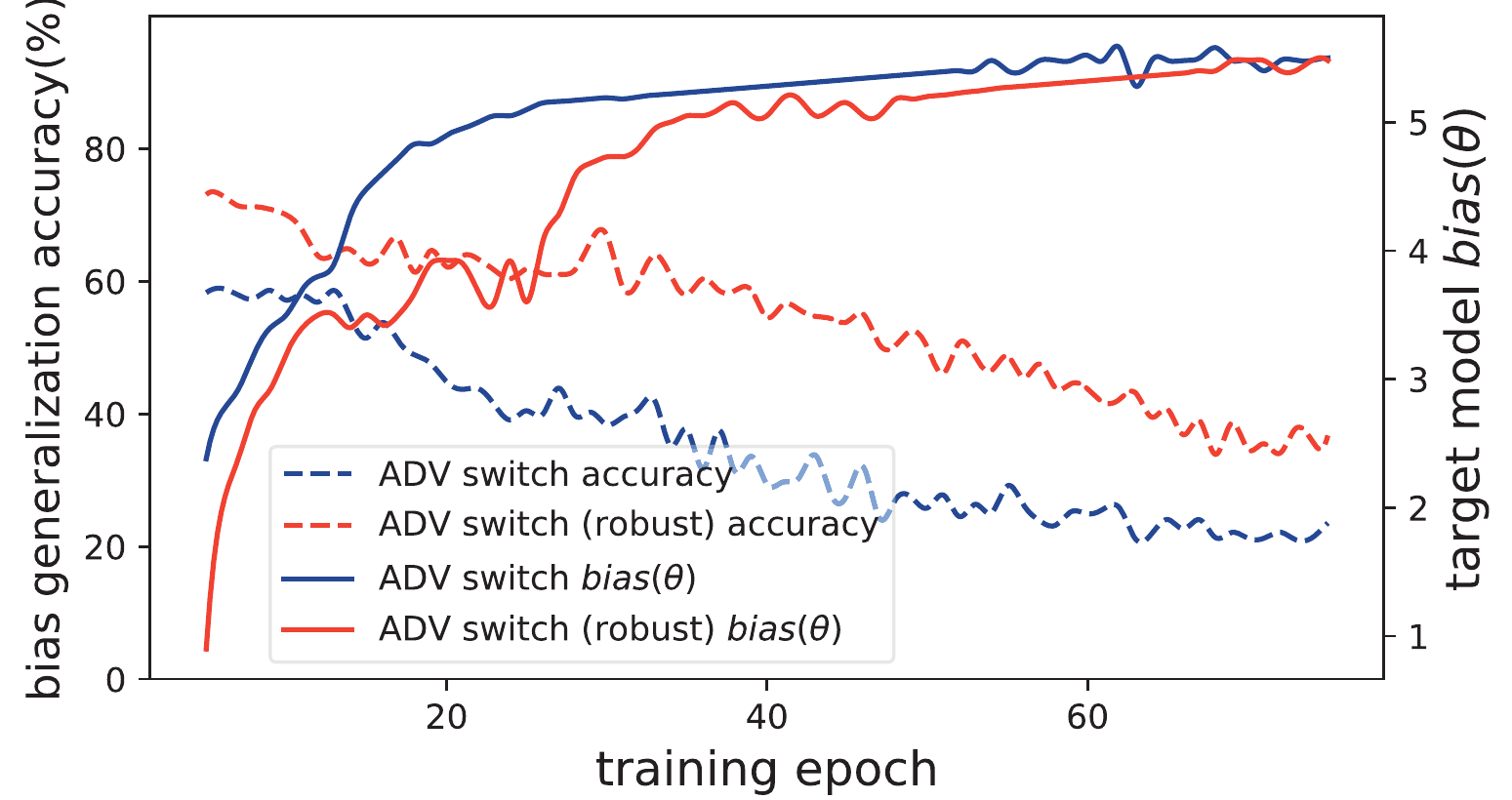}
  \caption{Cross-task transferability of adversarial examples during the training process of the target task classifier.}
  \label{fig4}
\end{figure}
To further track whether cross-task transferability is maintained during the training process, for every training epoch $m$, we repeated the following three operations: 
(1) using $\mathcal{X}_{augment}$ to update $f(\cdot)^{(m)}$ and $h_t(\cdot)^{(m)}$ for target task; (2) fixing $f(\cdot)^{(m)}$ and using $\mathcal{X}_{adv}$ with attacked gender labels to update $h_b(\cdot)_{generalize}^{(m)}$ for bias task; 
(3) using $\{f(\cdot)^{(m)};h_b(\cdot)_{generalize}^{(m)}\}$to test $\mathcal{X}_{org}$ and calculating the generalization accuracy $r^{(m)}$. 
Figure~\ref{fig4} illustrates the generalization accuracy (left $y-axis$) for every training epoch, where \emph{ADV switch} and \emph{ADV switch (robust)} correspond to the results by using $g_{ori}$ and $g_{robust}$ to generate adversarial examples respectively. 
It is shown the adversarial examples generated from $g_{ori}$ and $g_{robust}$ both gradually lose cross-task transferability as training proceeds. We explain the reason to yield this result as 
that: under the optimization goal of minimizing target task loss, the feature extractor $f(\cdot)$ tends to ignore adversarial information of adversarial examples, and the $\{f(\cdot)^{(m)};h_b(\cdot)_{generalize}^{(m)}\}$ trained on $\mathcal{X}_{adv}$ largely fail to classify original samples. 

To demonstrate the influence of cross-task transferability to model bias, we also calculated the gender bias of target task model during training, which is illustrated in Figure~\ref{fig4} with 
right y-axis. It is shown that model bias generally increases as adversarial examples lose their cross-task transferability. Combining with the above explanation for the decreased generalization accuracy, 
we understand the correlation between model bias and cross-task transferability as that: when $f(\cdot)$ tends to ignore useful adversarial information of adversarial examples, the role of 
augmenting adversarial examples reduces to replicating the original samples and derives trivial effect in balancing data distribution. Therefore, adjusting the generated adversarial examples to fit to the ever updating 
feature extractor and maintaining the cross-task transferability is critical for adversarial example-based data augmentation for visual debiasing. 

\begin{figure}[t]
  \centering
  \includegraphics[width=0.95\linewidth]{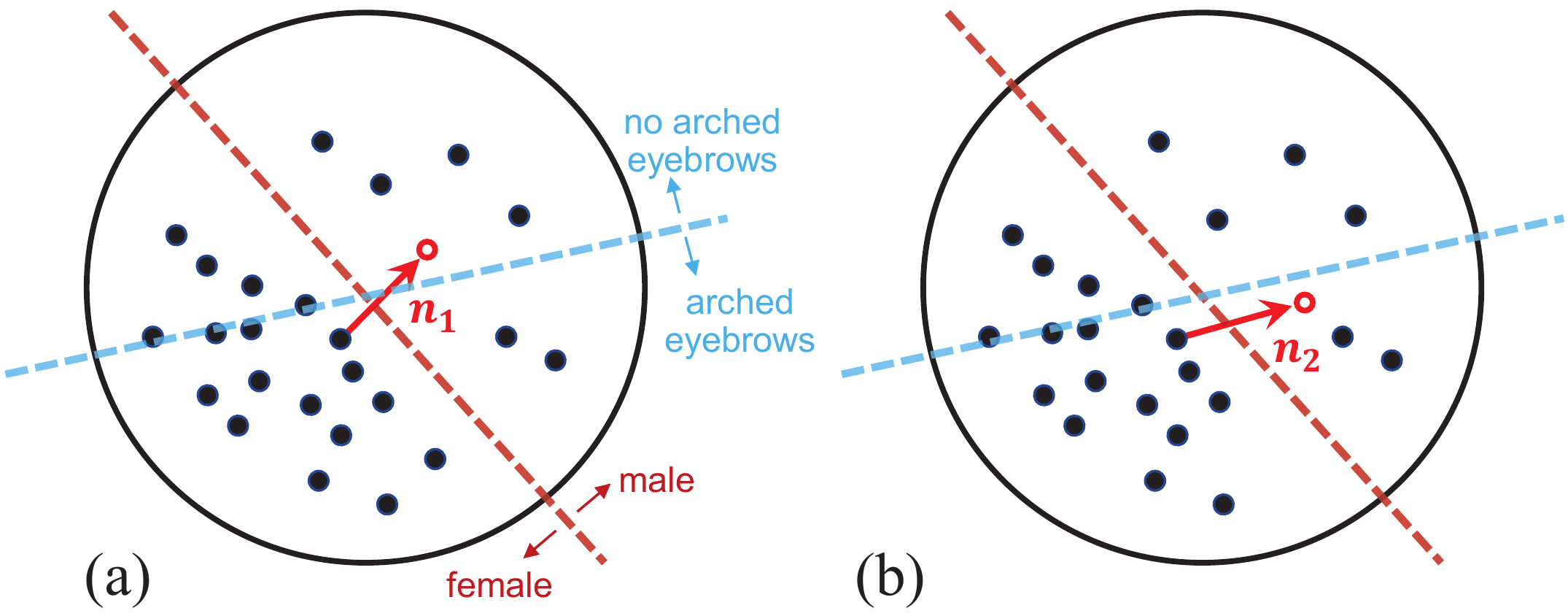}
  \caption{Illustration of adversarial perturbation in feature space: (a) only considering the bias task (e.g., female/male); (b) jointly considering the target task (e.g., ``arched eyebrow'' prediction).}
  \label{fig5}
\end{figure}

\begin{figure*}[t]
  \centering
  \includegraphics[width=0.92\linewidth]{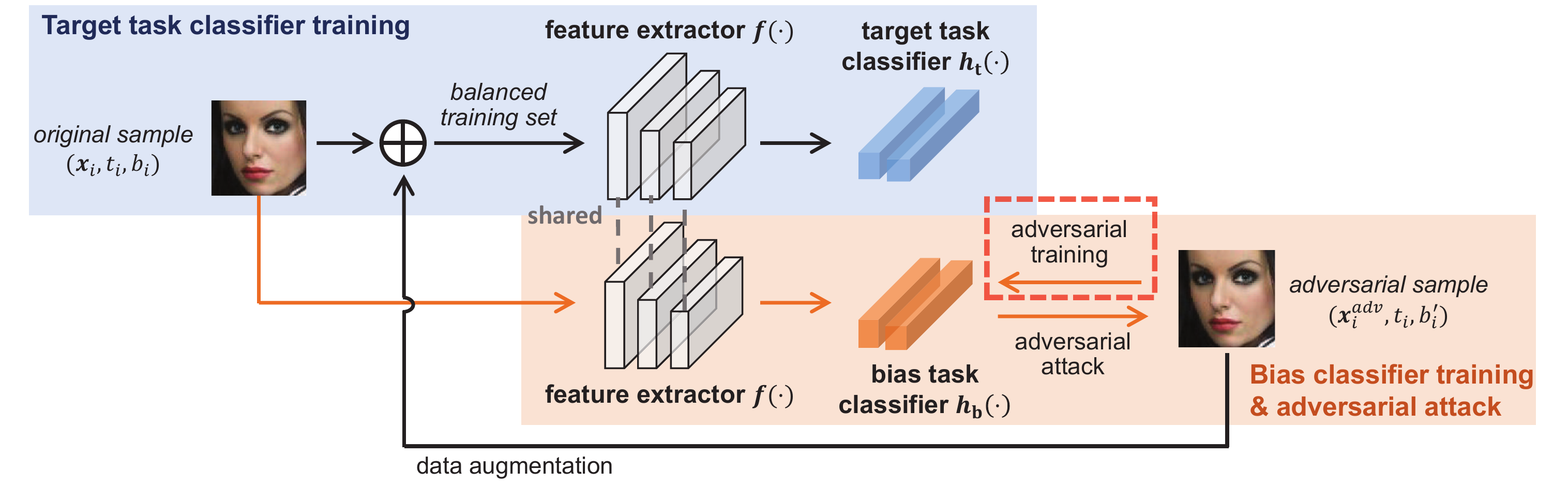}
  \caption{Framework of \emph{AEDA$\_$online} and \emph{AEDA$\_$robust}. \emph{AEDA$\_$robust} contains additional adversarial training module (highlighted with red dashline) to \emph{AEDA$\_$online}. }
   \label{fig-framework}
\end{figure*}

\section{Methodology}
\subsection{Overview}
We formally define visual debiasing problem as follows:

\textsc{Definition 2} (\textsc{Visual Debiasing}). \emph{Given image data set 
$\mathcal{D}_{ori}=\{\mathbf{x}_i,t_i,b_i\}_{i=1:N}$, where $\mathbf{x}_i\in \mathcal{X}_{ori}$ denotes the $i^{th}$ image original feature, $t_i\in \mathbb{T}$ denotes its target task label, $b_i\in \mathbb{B}$ denotes its 
bias task label, visual debiasing aims to learn an unbiased target task
classifier with parameter $\theta $ satisfying that the prediction of target task label $\hat{t}$ is 
independent of the bias labels: $P(\hat{t} = t_i|b,t^*;\theta) = P(\hat{t} = t_i|t^*;\theta)$. }

The above data analysis section justifies the attribution of model bias from imbalanced data distribution and the potential of adversarial example in balancing data distribution. Following these observations, in this section, we propose the visual debiasing solution via Adversarial Example-based Data Augmentation (AEDA). The direct way to realize \emph{AEDA} is to separately generate adversarial examples to balance data distribution as pre-processing and then use the augmented dataset to train target task classifier. This leads to the basic version of our solution, which we call \emph{AEDA$\_$pre} and will be introduced in Section~\ref{sec3.2}. To address the cross-task transferability issue, we propose to couple target task classifier training and adversarial example generation, which we call \emph{AEDA$\_$online} and will be introduced in Section~\ref{sec3.3}.  A complete version of our solution called \emph{AEDA$\_$robust} is elaborated in Section~\ref{sec3.4} to further address the adversarial generalization issue, where we conduct additional adversarial training operation when updating the bias classifier to improve the robustness.

\subsection{AEDA$\_$pre}\label{sec3.2}

In typical visual debiasing problems, among the training samples with specific target task label $t\in \mathbb{T}$, the ratio of samples with different bias labels $b\in \mathbb{B}$ is usually heavily imbalanced. 
For example, regarding the classification of ``arched eyebrows'' in CelebA, the ratio of samples with female gender to male gender is above $15:1$.  To balance the data distribution for visual debiasing, 
the direct way to realize \emph{AEDA} consists of two steps: (1) Bias variable adversarial attack, adding perturbation to the original sample $\mathbf{x}_i \in \mathcal{X}_{ori}$ to 
generate adversarial examples $\mathbf{x}_i^{adv}$ with the altered bias label $b_i^{'}$. In this way, adversarial attack supplements the shortage of minority-bias variable samples (e.g., male ``arched eyebrows'' images) to construct 
a balanced dataset ${\mathcal{D}_{augment}=\{\mathcal{D}_{ori};\{\mathbf{x}_i^{adv}, t_i, b_i^{'}\}\}}$. (2) Target task classifier training, using the balanced dataset $\mathcal{D}_{augment}$ to train an unbiased target task classifier.

Regarding the first step, using I-FGSM as the adversarial attack method, we alter the bias label for the $i^{th}$ image as follows:
\begin{equation}\label {eqn2}
\mathbf{x}_i^{adv}=\underset{\mathbf{x}_i}{\arg\min } \; L_{bias}\left(\mathbf{x}_i, b_{attack}\right)
\end{equation}
where $L_{bias}$ is the loss function of the bias classifier, $b_{attack}$ is the bias attack class. While standard adversarial attack changes the bias label of original samples, the added adversarial perturbation has risk to also affect the feature representation for target task prediction. As shown in Figure~\ref{fig5}(a), with the identified perturbation direction $\mathbf{n}_1$ by only considering the bias classifier gradients, adversarial perturbation is risky to also move across the target task classification boundary. This tends to deteriorate the training of target task classifier.

Therefore, we revise the adversarial attack step and require the generated adversarial examples to also possess the following property: the added adversarial perturbation should not affect the target task variable of the original sample. To implement this, we modify Eqn.~\eqref{eqn2} and generate adversarial examples as follows:

\begin{equation}\label {eqn3}
  \mathbf{x}_i^{adv}=\underset{\mathbf{x}_i}{\arg\min } \Big( \lambda L_{bias}(\mathbf{x}_i, b_{attack})+(1-\lambda)L_{target}(\mathbf{x}_i, t)\Big)
\end{equation}
where $L_{target}$ is the loss function of the target task classifier~\footnote{~\small{A preliminary target task classifier needs to be learned first from the original training set before adversarial data augmentation.}}, $t$ is target task label and $\lambda$ is the weighting parameter controlling the two terms. The added term plays role in preventing the target task features from damaged by adversarial perturbation and thus preserving the target task label.  As shown in Figure~\ref{fig5}(b), by jointly considering the target task classifier, the identified perturbation direction $\mathbf{n}_2$ is guaranteed to preserve the feature and label for target task prediction.

\subsection{AEDA$\_$online}\label{sec3.3}
As observed in Section~\ref{sec2.3} that adversarial examples have poor cross-task transferability when training target task classifiers. This will lead to a ``fake'' balanced data distribution, which contributes little to data augmentation and visual debiasing. To address this, instead of separating adversarial attack and target task classifier training as in \emph{AEDA$\_$pre}, we propose to couple these two steps by restricting the bias classifier utilizing the feature extractor of target task classifier. As shown in Figure~\ref{fig-framework}, target task and bias task share the same feature extractor $f(\cdot)$, with two additional classification modules $h_t(\cdot)$ and $h_b(\cdot)$ on top of $f(\cdot)$. By attacking the coupled bias classifier $\{f(\cdot);h_b(\cdot)\}$, the generated examples are expected to deliver discriminative features from $f(\cdot)$ and thus ensure cross-task transferability to the target task. 

However, this one-time coupled adversarial attack can only guarantee cross-task transferability for few epoches. The observation in Section~\ref{sec2.3} demonstrates that adversarial examples gradually lose their cross-task transferability and fail to continuously balance the data distribution during training of the target task classifier. To address this, we design online coupled adversarial attack where target task classifier and bias task classifier are simultaneously updated. Specifically, the following three steps iterate during the training process (at epoch $m$):
\begin{itemize}
\item Target task classifier training: 
\begin{equation}\label{eq-online1}
\min _{f, h_{t}}  L_{\text {target}}(\{\mathcal{X}_{ori},\{\mathcal{X}_{adv}^{(m-1)}\}, t) 
\end{equation}
\item Bias task classifier training: 
\begin{equation}\label{eq-online2}
\min_{h_{b}} L_{bias}(\mathcal{X}_{ori},b;f(\cdot))
\end{equation}
\item Adversarial attack: 
$\mathcal{X}_{adv}^{(m)}\leftarrow \mathcal{X}_{ori}$ following Eqn.~\eqref{eqn3}
\end{itemize}

In this way, the adversarial examples are adaptively generated based on the current target task feature extractor and thus guaranteed to well transfer to next epoch's target task training.

\subsection{AEDA$\_$robust}\label{sec3.4}
Section~\ref{sec2.2} observes that the capability of adversarial examples to generalize to attack class largely depends on the robustness of the attack classifier to generate the adversarial examples. To improve the adversarial generalization, based on \emph{AEDA$\_$online}, we further introduce an adversarial training module when updating the bias task classifier. 

As highlighted in red dashline of Figure~\ref{fig-framework}, the generated adversarial examples at previous epoch are employed in an adversarial training setting towards robust bias classifier. Specifically, the training process is modified to iterate the following three steps:
\begin{itemize}
\item Target task classifier training: same as Eqn.~\eqref{eq-online1};
\item Robust bias task classifier training (the adversarial samples are employed in training at intervals of $k$ mini-batches, and $1/k$ represents the frequency of adversarial training): 
\begin{equation}
\min_{h_{b}} L_{bias}(\{\mathcal{X}_{ori},\mathcal{X}_{adv}^{(m-1)}\},b;f(\cdot))
\end{equation}
\item Adversarial attack: 
$\mathcal{X}_{adv}^{(m)}\leftarrow \mathcal{X}_{ori}$ following Eqn.~\eqref{eqn3}
\end{itemize}

By iteratively optimizing between the above three steps, the generated adversarial examples are guaranteed to continuously hold the generalization capability to bias attack class as well as maintain the cross-task transferability to target task. The optimization ends when the training loss of target task converges. 

\section{Experiments}

\subsection{Experiment Setup}
We evaluated the proposed \emph{AEDA} solution on both simulated and real-world bias scenarios: 
(1) For the simulated bias, we used C-MNIST dataset~\cite{lu2018attribute}. Its 10 handwritten digit classes (0-9) were regarded as the target task variables, and the associated background colors were regarded as the bias variable. 
When classifying digit classes, the model may rely on the background color of image.
The goal is to remove the background color bias in digit recognition. (2) For the real-world bias, we used the face dataset CelebA with multiple facial attributes as the target task variables. 
The goal is to remove the model’s gender bias in facial attribute classification.

\noindent\textbf{Baselines}\hspace{2mm} We consider several typical debiasing solutions for comparison. 

\noindent\textbf{$\bullet$} Down-sampling: discarding samples with majority bias variable to construct a balanced dataset before target task training~\cite{drummond2003c4,zhou2005training}.

\noindent\textbf{$\bullet$} Reweighting: assigning different weights to samples and modifying the training objectives to softly balance the data distribution~\cite{kamiran2012data}.

\noindent\textbf{$\bullet$} Adv debiasing: the typical in-processing debiasing solution by adversarially learning between the target and bias tasks~\cite{wadsworth2018achieving,beutel2017data}. The goal is to extract fair representation that contributes to the target task but invalidates the bias task. We followed the previous studies and implemented Adv debiasing using transfer reversing gradient~\cite{ganin2014unsupervised}.

\noindent\textbf{$\bullet$} CycleGAN: generating synthetic data by altering bias variables~\cite{zhu2017unpaired}. We implemented this to compare \emph{AEDA} with the line of solutions explicitly supplementing minority samples.

\noindent\textbf{Evaluation metrics}\hspace{2mm} For target task performance, since the examined datasets involve with multi-label classification,  
we use balanced average accuracy (bACC) for evaluation: bACC is defined as the average accuracy of each group with different target variables and bias variables. 
From a fairness perspective, the importance of people with different bias variables is equal. An unbiased evaluation of accuracy should assign the same weight to groups with different bias variables. The higher the bACC, the more accurate the target task classifier is. 
For the model bias $bias(\theta)$, the lower the $bias(\theta)$, the more fair the model is.

\begin{figure}[t] 
  \centering
  \includegraphics[width=0.98\linewidth]{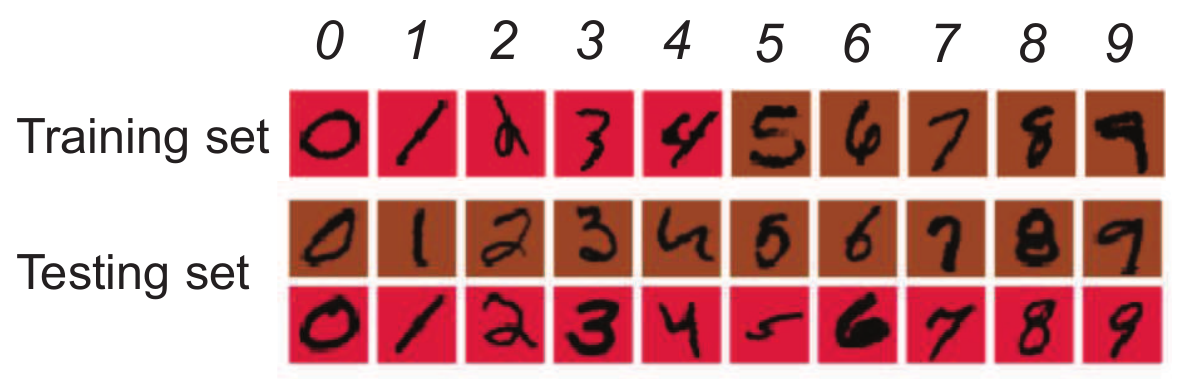}
  \setlength{\abovecaptionskip}{3pt}
  \setlength{\belowcaptionskip}{-4mm}
  \caption{The training and testing setting of simulated debiasing.}
  \label{fig7a}
\end{figure}

\subsection{Simulated Debiasing Evaluation}
We modified the C-MNIST data set for background color debiasing performance evaluation. 
In the training data, digits 0$\sim$4 are associated with red background color, 5$\sim$9 are associated with brown background color, 
and other data are dropped. The data with $\left \langle0\sim9, red\right \rangle$ and $\left \langle0\sim9, brown\right \rangle$ are both used for testing. 
The modified training and testing settings are illustrated in Figure~\ref{fig7a}. 
It is easy to see that the background color of the digits is the simulated bias variable, which contains high-predictive information but are actually independent of the target digit classes.  
With the testing set as a fair distribution setting, we used it to evaluate the target task accuracy and model bias. 


The debiasing methods are to make the model independent of the background color of the image in digit recognition. 
Table~\ref{tab1} summarizes the performance for different methods. It is shown that with the very different distributions between training and testing dataset, 
the \emph{Original} classifier achieves a rather low bACC at $55.62\%$. Moreover, since the modified C-MNIST training set is extremely imbalanced with only one background color for each digit, 
it is impossible to employ \emph{Down-sampling} and \emph{Reweighting}. Other main observations include: (1) all the proposed three settings of \emph{AEDA} obtain superior bACC and model bias than that of \emph{Original}. 
By simultaneously robustly training the bias classifier and coupling it with the target task classifier training, \emph{AEDA\_robust} demonstrates the best performance over all baselines in both bACC and model bias. 
(2) \emph{CycleGAN} obtains similar performance with \emph{AEDA\_pre}, showing very limited debiasing result due to the separation of pseudo sample generation and target task training. (3) \emph{Adv debiasing} obtains noticeable improvement in both bACC and model bias. We owe this result to the relative easy disentanglement between shape and color in the C-MNIST dataset, where \emph{Adv debiasing} manages to extract fair representation focusing on shape feature and only correlating to the target task.

\begin{table}
  \caption{Performances comparison on the C-MNIST data set. }
  \label{tab1}
  \begin{tabular}{l|c c}
    \toprule
    Methods&bACC (\%)&Model bias\\
    \midrule
    Original& 55.62& 7.84\\
    Under-sampling~\cite{drummond2003c4,zhou2005training}& --& --\\
    Reweighting~\cite{kamiran2012data}& --& --\\
    Adv debiasing~\cite{ganin2014unsupervised,wadsworth2018achieving,beutel2017data}& 89.93& 1.37\\
    CycleGAN~\cite{zhu2017unpaired}& 65.23& 5.42\\
    \midrule
    AEDA$\_$pre& 64.53& 5.89\\
    AEDA$\_$online& 80.57& 3.20\\
    AEDA$\_$robust& \textbf{91.80}& \textbf{0.53}\\
  \bottomrule
\end{tabular}
\end{table}

\begin{figure}[t] 
  \centering
  \includegraphics[width=0.8\linewidth]{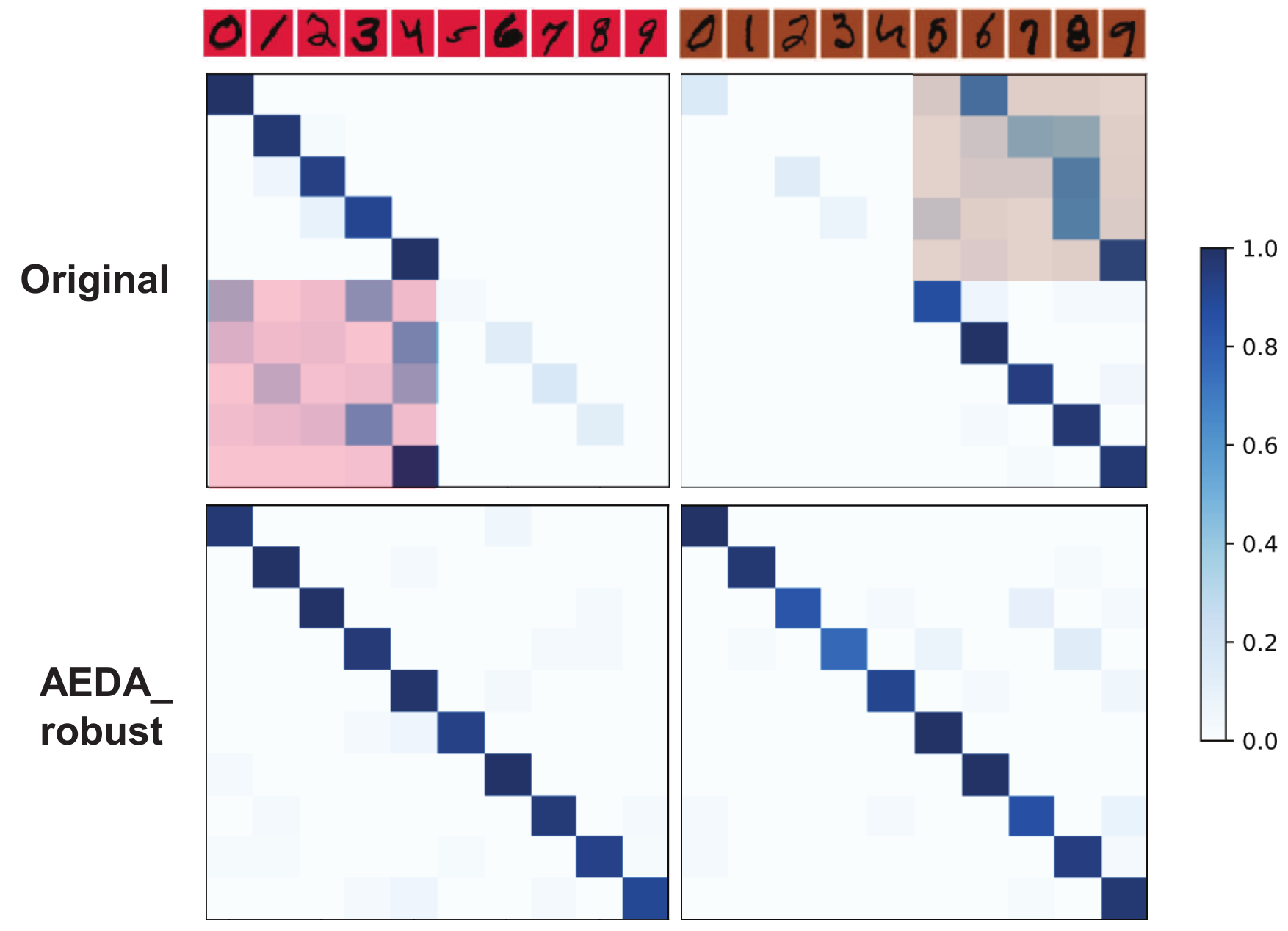}
  \setlength{\belowcaptionskip}{-3mm}
  \caption{Digit classification confusion matrices for testing subset of $\left \langle0\sim9, red\right \rangle$ (left) and $\left \langle0\sim9, brown\right \rangle$ (right).}
  \label{fig7b}
\end{figure}

\begin{figure}[h] 
  \centering
  \includegraphics[width=0.99\linewidth]{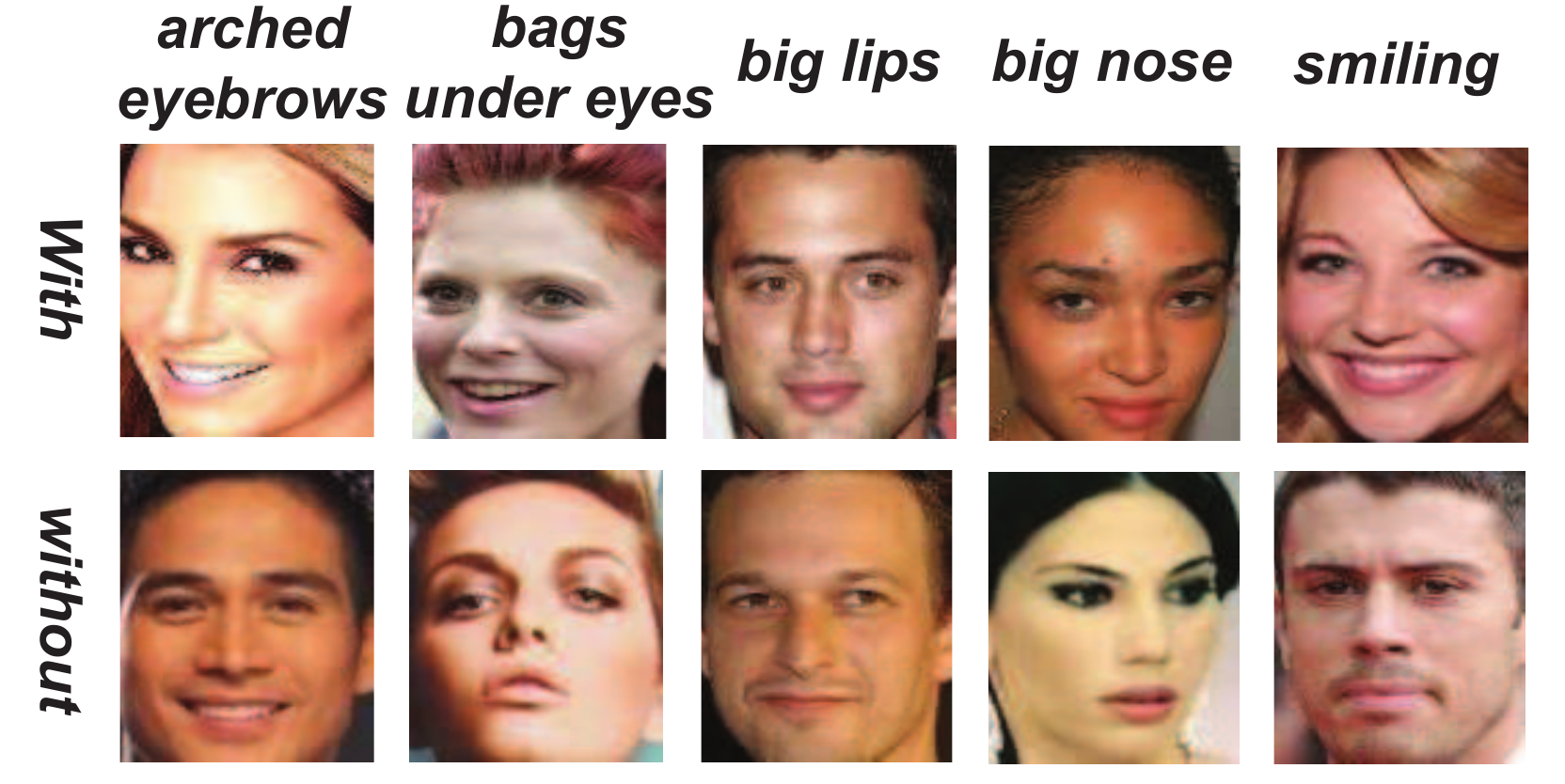}
  \setlength{\abovecaptionskip}{-3pt}
  \setlength{\belowcaptionskip}{-5mm}
  \caption{Example images of 5 facial attributes to be predicted. The CelebA data set also provides gender annotation for each image.}
  \label{fig8a}
\end{figure}

To further evaluate the influence of training distribution on model performance for different target task variables, we examined the classification accuracy for each digit class on the $\left \langle0\sim9, red\right \rangle$ and $\left \langle0\sim9, brown\right \rangle$ testing subsets respectively. 
The results obtained by \emph{Original} and the proposed \emph{AEDA\_robust} are summarized as confusion matrices in Figure~\ref{fig7b}. Observations include: (1) The classification accuracy of \emph{Original} is severely dependent on the background color of the testing subset, while \emph{AEDA\_robust} obtains consistent classification performance between the two subsets. This coincides with the definition of model bias in Eqn.~\eqref{eqn1} and validates the model debiasing effectiveness. (2) For \emph{Original} classifier, when the training and testing set have very different data distributions (e.g., bottom-left for $\left \langle0\sim9, red\right \rangle$ subset and upper-right for $\left \langle0\sim9, brown\right \rangle$ subset), the testing accuracy tends to be very low. This also demonstrates that the imbalanced training data distribution heavily influences not only the model bias but also the target task performance.

\begin{figure*}[t] 
  \centering
  \includegraphics[width=0.97\linewidth]{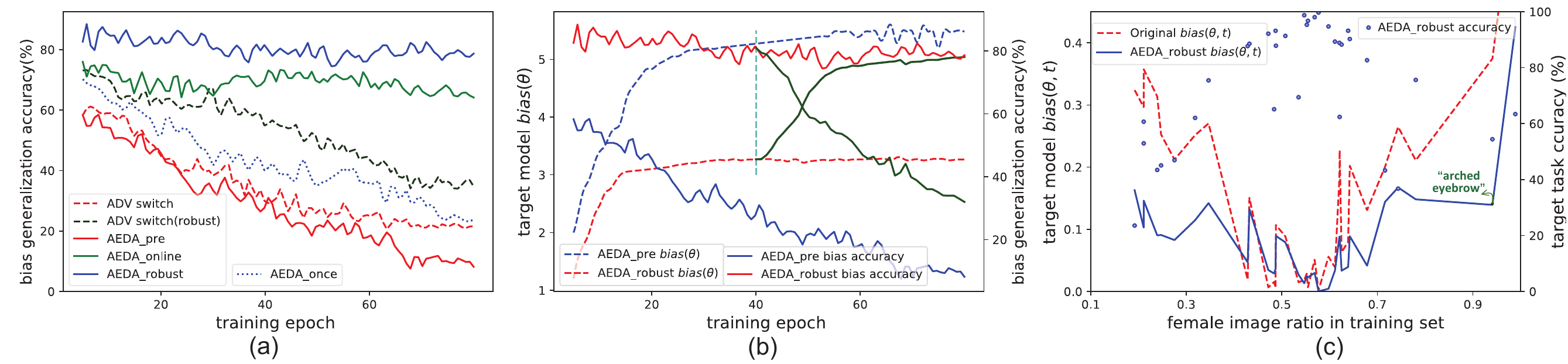}
  \setlength{\abovecaptionskip}{1pt}
  \setlength{\belowcaptionskip}{-7pt}
  \caption{(a) Cross-task transferability during training the target task.
  (b) Model bias and cross-task transferability in \emph{AEDA\_pre} and \emph{AEDA\_robust}. 
  (c) The debiasing effectiveness and traget task accuracy for each examined facial attribute.}
  \label{fig9}
\end{figure*}

\begin{table}[t]
  \caption{Performances comparison on the CelebA data set.}
  \label{tab2}
  \begin{tabular}{l|c c}
    \toprule
    Methods&bACC (\%)&Model bias\\
    \midrule
    Original& 73.57& 5.48\\
    Under-sampling~\cite{drummond2003c4,zhou2005training}& 66.35& \textbf{2.35}\\
    Reweighting~\cite{kamiran2012data}& 73.82& 4.39\\
    Adv debiasing~\cite{ganin2014unsupervised,wadsworth2018achieving,beutel2017data}& 72.82& 4.23\\
    CycleGAN~\cite{zhu2017unpaired}& 73.65& 4.75\\
    \midrule
    AEDA$\_$pre& 73.68& 5.23\\
    AEDA$\_$online& 74.03& 4.22\\
    AEDA$\_$robust& \textbf{74.30}& 3.27\\
  \bottomrule
\end{tabular}
\end{table}
\subsection{Real-world Debiasing Evaluation}
We evaluated the real-world gender debiasing performance on the CelebA data set and used VGG-16~\cite{simonyan2014very} as the backbone network. The same 34 facial attributes are employed as the target tasks. Example images of 5 facial attributes to be predicted are illustrated in Figure~\ref{fig8a}. Moreover, in the original CelebA, different images of the same person may appear in both training and testing set. To prevent person-specific features contributing to facial attribute prediction, we keep one image for each person in either training or testing set. 

Table~\ref{tab2} summarizes the performance for different methods on CelebA. All reported results are averaged from 5-fold cross validation. The consistent observations with the above simulated debiasing evaluation include: (1) Among the the three \emph{AEDA} settings, \emph{AEDA\_robust} performs best in both bACC and model bias. (2) \emph{CycleGAN} demonstrates limited debiasing performance. New observations include: (1) \emph{Original} has normal bACC at $73.57\%$. In this case, instead of sacrificing accuracy for fairness, it is interesting to see that the proposed \emph{AEDA} solutions still obtain superior performance than \emph{Original} in both accuracy and fairness. In the next subsection, we will discuss further the role of data augmentation-based solution in improving accuracy as well as fairness. (2) \emph{Down-sampling} and \emph{Reweighting} are valid to reduce the model bias for facial attribute prediction, with \emph{Down-sampling} obtaining remarkable debiasing performance by explicitly balancing data distribution. However, by discarding samples with majority-bias variable and employing a fragmented training set, \emph{Down-sampling} achieves a very poor target task performance (bACC=$66.35\%$). (3) \emph{Adv debiasing} shows certain debiasing performance, but not as good as that in the simulated evaluation. This difference is due to the complicated coupling between the target and bias variables in real-world debiasing scenarios.

\noindent\textbf{The necessity of online coupled adversarial attack.}\hspace{2mm}  
To further understand the mechanism of online coupled adversarial attack, we reported more results by examining the training process. Recalling from Section~\ref{sec2.3}, cross-task transferability is critical for debiasing. Therefore, we first calculated the bias generalization accuracy $r^{(m)}$ at each training epoch for the proposed \emph{AEDA} solutions, which is shown in Figure~\ref{fig9}(a). While \emph{AEDA\_pre} shows rather poor bias generalization accuracy, \emph{AEDA\_online} and \emph{AEDA\_robust} demonstrate stable and superior capability in maintaining cross-task transferability. To analyze the role of online continuous attack, a fourth setting named \emph{AEDA\_once} is also implemented and compared, which is similar to \emph{AEDA\_online} except with the difference that adversarial examples $\mathcal{X}_{adv}$ are generated only once instead of continuously updated during the training process. As shown in Figure~\ref{fig9}(a), although better than \emph{AEDA\_pre}, the cross-task transferability of \emph{AEDA\_once} fails to retain as training proceeds, validating the necessity for online coupled adversarial attack.

We then analyzed the influence of cross-task transferability to model debiasing. Using \emph{AEDA\_pre} and \emph{AEDA\_robust} as examples, we examined both the bias generalization accuracy (right y-axis) and the target model bias (left y-axis) in Figure~\ref{fig9}(b). After around 15 training epochs, it is shown that the model bias of \emph{AEDA\_robust} is stable at a relative low level ($3.1$). However, the model bias of \emph{AEDA\_pre} continues to increase as its cross-task transferability gradually loses, with a final model bias as high as $5.3$. Moreover, we modified the training process of \emph{AEDA\_robust} and examined the changes of cross-task transferability and model bias when terminating online update of adversarial example generation at $40^{th}$ epoch. Shown with green curves in Figure~\ref{fig9}(b), the cross-task transferability and model bias approach those of \emph{AEDA\_pre} after $40^{th}$ epoch, further demonstrating the effectiveness of online coupled adversarial attack in retaining cross-task transferability and reducing model bias. 

\subsection{Discussion}
\subsubsection{The consistency between accuracy and fairness}
In addition to alleviating model bias, the proposed \emph{AEDA} solution gives rise to two interesting discussions. First discussion involves with the accuracy-fairness paradox. It is recognized in conventional debiasing attempts ~\cite{zhao2019inherent,fish2016confidence,menon2017cost,feldman2015certifying} that, there exists tradeoff between accuracy and fairness and the goal is to reduce model bias under the condition of equal, if not slightly decreased accuracy. 

However, the results in both simulated and real-world debiasing evaluation demonstrate \emph{AEDA}'s capability in simultaneously improving model accuracy and fairness. In Figure~\ref{fig9}(c), we show the model bias and target task accuracy for each examined facial attribute. Other than the consistently reduced model bias from \emph{Original}, it is interesting to find the close relation between distribution balance, model bias and accuracy: the facial attributes with more balanced data distribution tend to have lower model bias and higher accuracy. 

This inspires us to examine the common attributions of accuracy and fairness to data distribution. By generating supplementary samples to augment the training data, 
a balanced dataset contributes to removing the erroneous dependence on bias variables and thus making the inference of target tasks more dependent on intrinsic features. Without discarding training data or adding constraints to affecting target task learning, the proposed data augmentation-based solution provides alternative perspective to address model debiasing problem and validates the possibility to simultaneously improving fairness and accuracy. 
\subsubsection{Adversarial attack-based pseudo sample generation}
The previous study has discovered that human and model may rely on very different features~\cite{ilyas2019adversarial}. In the above experiments, we verified this in the context of model debiasing tasks that model can extract useful information from training samples that human hardly recognizes. In most supervised learning scenarios, the main challenge is to collect adequate samples recognizable from human's perspective. Without the restrictions of human recognition, adversarial attack actually provides alternative solution to generate pseudo samples and makes up for the common data shortage issues.

In this spirit, we examined the potential of adversarial attack-generated samples in few shot learning. Specifically, the proposed \emph{AEDA\_robust} is implemented and 4 handwritten digit classes as few shot classes experiment is conducted in the C-MNIST settings. A $27\%$ average improvement in few shot classes is obtained than \emph{Original}, showing the feasibility of adversarial attack-generated pseudo sample in addressing problems beyond debiasing. Moreover, adversarial attack enjoys advantages that human prior knowledge is readily to be incorporated to generate pseudo samples with desired properties to increase sample diversity.

\section{Conclusion}
In this work, we propose to balance data distribution via adding supplementary adversarial example towards visual debiasing. The proposed solution couples the operation of target task model training, bias task model training, and adversarial sample generation in an online learning fashion. The experimental results in both simulated and real-world debiasing evaluation demonstrates its effectiveness in consistently improving fairness and accuracy. In addition to study the common attribution behind fairness and accuracy, it will also be interesting to employ adversarial example to assist learning tasks other than model debiasing in the future. 

\begin{acks}
This work is supported by the National Key R\&D Program of China (Grant No. 2018AAA0100604), and the National Natural Science Foundation of China (Grant No. 61632004, 61832002, 61672518).
\end{acks}


\bibliographystyle{ACM-Reference-Format}
\balance
\bibliography{ref}

\end{document}